# Robust Independence Testing for Constraint-Based Learning of Causal Structure

Denver Dash
Decision Systems Laboratory
Intelligent Systems Program
University of Pittsburgh
Pittsburgh, PA 15260

Marek J. Druzdzel
Decision Systems Laboratory
Intelligent Systems Program and
School of Information Sciences
University of Pittsburgh
Pittsburgh, PA 15260

## Abstract

This paper considers a method that combines ideas from Bayesian learning, Bayesian network inference, and classical hypothesis testing to produce a more reliable and robust test of independence for constraint-based (CB) learning of causal structure. Our method produces a smoothed contingency table $\overline{N}_{XYZ}$ that can be used with any test of independence that relies on contingency table statistics. $\overline{N}_{XYZ}$ can be calculated in the same asymptotic time and space required to calculate a standard contingency table, allows the specification of a prior distribution over parameters, and can be calculated when the database is incomplete. We provide theoretical justification for the procedure, and with synthetic data we demonstrate its benefits empirically over both a CB algorithm using the standard contingency table, and over a greedy Bayesian algorithm. We show that, even when used with noninformative priors, it results in better recovery of structural features and it produces networks with smaller KL-Divergence, especially as the number of nodes increases or the number of records decreases. Another benefit is the dramatic reduction in the probability that a CB algorithm will stall during the search, providing a remedy for an annoying problem plaguing CB learning when the database is small.

## 1 Introduction

Constraint-based (CB) causal discovery searches a database for independence relations and constructs graphical structures called "patterns" which represent a class of statistically indistinguishable directed acyclic graphs (DAGs). This method contrasts to those based on Bayesian concepts, which typically reduce to a search-and-score procedure on the space of DAGs.

Both CB and Bayesian approaches have advantages and disadvantages [9]. Constraint-based approaches are relatively quick, deterministic, and have a well-defined stopping criterion; however, they rely on an arbitrary significance level to test for independence, and they can be unstable in the sense that an error early on in the search can have a cascading effect that causes many errors to be present in the final graph [13, 4]. Both approaches have the ability to incorporate background knowledge in the form of temporal ordering, or forbidden or forced arcs, but Bayesian approaches have the added advantage of being able to flexibly incorporate users' background knowledge in the form of prior probabilities over the structures and over the parameters of the network. Bayesian approaches are capable of dealing with incomplete records in the database by incorporating Monte Carlo sampling or greedy hill-climbing approaches such as the EM algorithm. The most serious drawback to the Bayesian approaches is the fact that they require a Markov chain search over structures and thus can be slow to converge.

When data sets are small, the relative benefits of the two approaches are unclear. One one hand, Bayesian methods seem to have an advantage because they can accommodate prior distributions which have a smoothing effect on the distributions in the sparse-data limit, on the other hand, using independence information can help restrict the massive search space [6].

Several approaches have attempted to mix the benefits of CB and Bayesian learning. Researchers have investigated performing greedy Bayesian searches over the space of equivalence classes of DAGs [11, 1]. More recently, Kočka and Castelo [10] investigated searching over the space of DAGs by using search operators that consider the search boundary to be defined by the equivalence class.

Researchers have also developed two-stage hybrid al-gorithms, where the first stage performs a constraint-based search and uses the resulting graph as input into a second-stage Bayesian search. In particular, [12] use the PC algorithm to generate an absolute temporal ordering on the nodes for use with the K2 algorithm [3], which requires such an ordering on the input. Spirtes and Meek [14] use the PC algorithm to generate a good starting graph for use in their greedy search over the space of essential graphs. Dash and Druzdzel [4] took the opposite approach, using an instability in CB learning to search the space of patterns, scoring each pattern using a Bayesian score. Friedman et al. [6] use independence information to restrict the search space of a greedy Bayesian algorithm.

Cooper [2], suggested a Bayesian independence test as part of an approximate constraint-based learning algorithm. This Bayesian test could evaluate conditional independence relations $(X \perp Y \mid Z)$, where $|Z| = 1$, by constructing various network fragments, scoring them according to a Bayesian score, and combining each score together into a single measure for independence. This approach is the closest to the one we present in this paper. Our technique differs in how the fragments are constructed, allowing arbitrary significance tests.

Our contributions in this paper are the following: (1) We develop a method by which it is possible to calculate a smoothed contingency table $\overline{N}_{XYZ}$ that can be calculated with missing data, allows the specification of priors, and can be calculated in the same time and with the same space requirements as a standard contingency table. (2) We provide theoretical justification for the use of $\overline{N}_{XYZ}$, (3) we demonstrate empirically that $\overline{N}_{XYZ}$ improves CB learning when using the chi-squared test $chi\text{-}squared(X, Y \mid Z, \mathbf{N}_{XYZ})$, and (4) we demonstrate empirically that when the number of records is small or the number of nodes is large, a CB algorithm can outperform a Bayesian greedy search over the space of DAGs.

In Section 2 we formally frame the problem, state our assumptions and notation, and review constraint-based learning techniques. In Section 3 we introduce a hybrid independence test, $Hybrid\text{-}IT(X, Y \mid \mathbf{Z}, D)$, that uses $\overline{N}_{XYZ}$ together with any standard test of independence $Std\text{-}IT(X, Y \mid \mathbf{Z}, D, \alpha)$, and prove the soundness and completeness of $Hybrid\text{-}IT(X, Y \mid \mathbf{Z}, D)$. In Section 4 we present experimental results, and in Section 5 we present our conclusions and future directions.

## 2 Learning Causal Models

Throughout this paper we use the notation $\mathbf{V} = \{V_1, V_2, \ldots, V_N\}$ to denote a set of $N$ random variables, and $D(\mathbf{V})$ to denote a database of $N_r$ records containing specific instances of the vector $\mathbf{V}$: $D = \{\mathbf{V} = \mathbf{v}_1, \ldots, \mathbf{V} = \mathbf{v}_{N_r}\}$. In general we use boldface notation to denote a set of objects and non-bold to denote singletons, when possible. We use uppercase symbols to denote random variables and lowercase symbols to denote specific values taken on by a random variable, e.g., $V_i = v$ or $\mathbf{V} = \mathbf{v}$. We use $\mathbf{Rng}(X)$ to denote the range of a variable $X$.

We use the notation $(X \perp Y \mid \mathbf{Z})$ to denote the fact that $X$ is independent from $Y$ given the set $\mathbf{Z}$.

**Assumption 1 (Multinomial variables)** *We assume that each node $X_i$ is a discrete variable with $r_i$ possible states.*

We let $r_{max}$ denote the maximum number of states: $r_{max} = \max_i r_i$. We let $v_i^k$ indicate the $k$-th state of variable $V_i \in \mathbf{V}$: $\mathbf{Rng}(V_i) = \{v_i^1, v_i^2, \ldots, v_i^{r_i}\}$.

A directed graph $G(\mathbf{V})$ is defined as a pair $\langle \mathbf{V}, \mathbf{E} \rangle$, where $\mathbf{E}$ is a set of directed edges $V_i \to V_j$, $V_i, V_j \in \mathbf{V}$. We use the notation $\mathbf{P}_i^G$ to denote the parent set of $V_i$ in $G$; however if $G$ is clear by the context we will drop the superscript. We use $\mathbf{p}_i^j$ to denote the $j$-th configuration of the parents of $V_i$: $\mathbf{P}_i \in \{\mathbf{p}_i^1, \ldots, \mathbf{p}_i^{q_i}\}$.

**Definition 1 (Bayesian network)** *A Bayesian network model $M$ over a set of variables $\mathbf{V} = \{V_1, \ldots, V_N\}$ is a pair $\langle G, \theta \rangle$, where $G(\mathbf{V})$ is a DAG over $\mathbf{V}$ and $\theta$ are a set of conditional probabilities: $\theta = \{\theta_{ijk} : \forall (ijk)\}$ such that $\theta_{ijk} = P(X_i = x_i^k \mid \mathbf{P}_i = p_i^j)$.*

In general we use the common $(ijk)$ coordinates notation to identify the $k$-th state and the $j$-th column of the $i$-th node in the network. In a causal model $M(\mathbf{V})$, we use the symbol $\theta_{ij}$ to denote the entire probability distribution function for the $i$-th node and the $j$-th column, and the symbol $\theta$ to denote the collective parameters of the network. We use "causal model" interchangeably with "Bayesian network".

Once a graphical structure $G$ has been constructed a Bayesian estimate for the parameters $\theta$ can be calculated in closed-form, given a few standard assumptions.

**Assumption 2 (Dirichlet priors)** *The prior beliefs over parameter values are given by a Dirichlet distribution.*

We let $N_{ijk}$ denote the number of times in the database that the node $X_i$ achieved state $k$ when $\mathbf{P}_i$ was in the $j$-th configuration, and we let $\alpha_{ijk}$ denote the Dirichlet hyperparameters corresponding to the network parameter $\theta_{ijk}$. We assume the hyperparameters $\alpha_{ijk}$ are given or can be calculated in $O(1)$ time. For example the non-informative K2 criterion [3] $\alpha_{ijk} = 1$ for all $(ijk)$ will satisfy this requirement.

**Assumption 3 (Parameter independence)** *For any given network structure S, each probability distribution $\theta_{ij}$ is independent of any other probability distribution $\theta_{i'j'}$:*

$$P(\theta \mid S) = \prod_{i=0}^{N} \prod_{j=1}^{q_i} P(\theta_{ij} \mid S) \quad (1)$$

Given the assumption of parameter independence and Dirichlet priors, it can be shown that a single network with a fixed set of parameters $\hat{\theta}$ given by

$$\hat{\theta}_{ijk} = \frac{\alpha_{ijk} + N_{ijk}}{\alpha_{ij} + N_{ij}} \quad (2)$$

will produce predictions equivalent to those obtained by averaging over all parameter configurations. [7] argued that under the canonical coordinate system these parameters represent a maximum *a posteriori* (MAP) configuration.

CB learning methods systematically check the data for independence relations and use those relationships to infer necessary features of the structure. The specific algorithm that we use in our experiments is a variant of the PC algorithm, and the reader is referred to [13] for details and proofs of correctness of the procedure. The main idea is presented below as a sketch.

We assume the existence of a standard independence test $Std\text{-}IT(X, Y \mid \mathbf{Z}, D, \alpha)$, a set of variables $\mathbf{V}$, a complete database $D$, and a significance level $\alpha$. The algorithm is sketched as follows:

**Procedure 1 (PC algorithm)**
Given: $\mathbf{V}$, $D$ and $\alpha$.

1. $S_u = Find\text{-}Independence\text{-}Graph(\mathbf{V}, D, \alpha)$,
2. $S = Orient\text{-}Edges(S_u, D)$,
3. Return $S$.

$Find\text{-}Independence\text{-}Graph(\mathbf{V}, D, \alpha)$ takes a set of variables $\mathbf{V}$ and a database $D$ as input and outputs an undirected graph $S_u$ such that an edge $X$—$Y$ exists in $S_u$ iff there does not exist a subset $\mathbf{Z} \subseteq \mathbf{V} \setminus \{X, Y\}$ (including the empty set) such that $Std\text{-}IT(X, Y \mid \mathbf{Z}, D) = true$. $S_u$ is constructed by checking conditional independence relations and removing edges from an initially complete undirected graph whenever an independence is found. The PC algorithm makes this procedure efficient by successively checking higher-order dependencies while restricting the set of nodes that need to be conditioned on. Specifically, let $Adj(A)$ denote the set of variables that are adjacent to $A$, then $Find\text{-}Independence\text{-}Graph(\mathbf{V}, D, \alpha)$ can be sketched as follows:

**Procedure 2 ($Find\text{-}Independence\text{-}Graph(\mathbf{V}, D, \alpha)$)**
1. Let $n = 0$.
2. Let $S_u$ be a complete undirected graph.
3. Repeat:

    (a) For all pairs of variables $(X, Y)$, check $Std\text{-}IT(X, Y \mid \mathbf{Z}, D, \alpha)$ for all subsets $\mathbf{Z}$ such that $|\mathbf{Z}| = n$ and $\mathbf{Z} \subset Adj(X)$ or $\mathbf{Z} \subset Adj(Y)$. If there exists a $\mathbf{Z}$ such that $Std\text{-}IT(X, Y \mid \mathbf{Z}, D, \alpha) = true$ then remove the edge $X$—$Y$ from $S_u$.

    (b) Set $n = n + 1$

    *Until no variable has greater than $n$ adjacencies, or a stopping condition is satisfied.*

4. Return $S_u$.

The sub-procedure $Orient\text{-}Edges(S_u, D)$ infers directionality of some arcs in $S$ by searching for independence relations characteristic of v-structures and by avoiding cycles. We use a modified version of PC that disallows cycles and bi-directed arcs. This modification was justified by the fact that our generating networks were acyclic with no latent variables, so if PC inferred such structures it must have been due to an error in some hypothesis tests during the search. This modification also makes comparison of the resulting structures much easier.

The graphs produced by CI-based procedures are partially directed graphs which go by several names in the literature, of which "pattern" and "essential graph" are the most common. Patterns summarize the structure of a Bayesian network that can be inferred from a list of independencies alone.

CB methods have the advantage of possessing clear stopping criteria and deterministic, systematic search procedures. On the other hand they are subject to several instabilities: namely, if a mistake is made early on in the search, it can lead to incorrect sets $Adj(A)$ and $Adj(B)$ later in the search which may in turn lead to bad decisions in the future, which can lead to even more incorrect sets $Adj(A)$, etc. This instability has the potential to cascade, creating many errors in the learned graph. Similarly, incorrect edges in $S_u$ can lead to incorrectly oriented arcs in the final graph $S$. It is for these reasons that the quality and reliability of the independence test is critical for practical constraint-based algorithms. Another disadvantage of CB methods is the difficulty of performing a classical (non-Bayesian) independence test when some data is missing from the database.

## 3 A Hybrid Independence Test

Standard statistical tests of independence, such as the chi-squared test, *chi-squared*$(X, Y \mid \mathbf{Z}, \mathbf{N}_{XY\mathbf{Z}})$, typically require the calculation of a set of statistics:

$$\mathbf{N}_{XY\mathbf{Z}} = \{N_{xy\mathbf{z}} : \forall\, x \in Rng(X),\, y \in Rng(Y),\, \mathbf{z} \in Rng(\mathbf{Z})\},$$

where $N_{xy\mathbf{z}}$ denotes the number of times that variable $X$ took value $x$, $Y$ took value $y$ and $\mathbf{Z}$ took configuration $\mathbf{z}$ in the database.

In this section we present a pseudo-Bayesian test of independence that uses Bayesian network inference together with a standard hypothesis test to perform tests of conditional independence. Our technique makes use of the fact that given a probability distribution $P(\mathbf{V})$, it is possible to calculate the expectation of $N_{XY\mathbf{Z}}$ for a database of size $N$ as:

$$\langle N_{xy\mathbf{z}} \rangle = N \cdot P(x, y, \mathbf{z}). \qquad (3)$$

For a Bayesian network $B(\mathbf{V}')$ over the set of variables $\mathbf{V}' = \{X, Y\} \cup \mathbf{Z}$, Equation 3 can be calculated in $O(|\mathbf{V}'|)$ time. When $\mathbf{V}'$ includes variables other than *set* $X, Y \cup \mathbf{Z}$, however, this calculation is less trivial because one cannot make full use of the ability to factorize the network; thus requiring in the worst case marginalization over the set of variables $\mathbf{V} \setminus \{\{X, Y\} \cup \mathbf{Z}\}$. Even if this calculation were feasible to perform, it would not in general be possible to use Bayesian network inference to estimate a contingency table in the inner loop of a CB discovery algorithm because we don't *know* the structure of the Bayesian network.

We propose a framework whereby network fragments containing only the variables relevant to the particular independence calculation are passed to a pseudo-Bayesian independence tester. We thus make use of the function $CalcStats(N, B)$ which takes a total number of records $N$ and a Bayesian network $B$ as input and outputs the expected statistics $\langle \mathbf{N}_{XY\mathbf{Z}} \rangle$ according to Equation 3:

**Procedure 3 ($CalcStats(X, Y, \mathbf{Z}, N, B)$)**
Given: $X$, $Y$, $\mathbf{Z}$, $N$ and $B$.

1. Calculate $P(x, y, \mathbf{z})$ for all $x \in Rng(X)$, $y \in Rng(Y)$ and $\mathbf{z} \in Rng(\mathbf{Z})$ using forward traversal over the network $B$.

2. Let $N_{xy\mathbf{z}} = N \cdot P(x, y, \mathbf{z})$ for all $x$, $y$ and $\mathbf{z}$.

3. return $N_{xy\mathbf{z}}$.

The complexity of Step 3 is $O(|\mathbf{Z}| \cdot |\mathbf{N}_{XY\mathbf{Z}}|)$ since $D_f$ includes only the variables $\{X, Y\} \cup \mathbf{Z}$.

We also assume the existence of a standard independence test $Std\text{-}IT(X, Y \mid \mathbf{Z}, \mathbf{N}_{XY\mathbf{Z}}, \alpha)$ which returns *true* or *false* based on the statistical threshold $\alpha$. The hybrid independence test is defined as follows:

**Procedure 4 ($Hybrid\text{-}IT(X, Y \mid \mathbf{Z}, D)$)**
Given: $D$, $X$, $Y$, $\mathbf{Z}$, and $\alpha$.

1. Construct a DAG fragment $D_f$ over the set $\{X, Y\} \cup \mathbf{Z}$ as follows: direct edges from $X \to Y$ and from $Z_i \to Y$ for all $Z_i \in \mathbf{Z}$.

2. Define the BN model $B_f \equiv \langle D_f, \theta_f \rangle$, where $\theta_f$ are given by Equation 2, and let $N_r \equiv |D|$.

3. Let $\overline{N}_{XY\mathbf{Z}} \equiv CalcStats(\langle X, Y, \mathbf{Z}, N_r, B_f \rangle)$,

4. Return $Std\text{-}IT(X, Y \mid \mathbf{Z}, \langle \overline{N}_{XY\mathbf{Z}} \rangle, \alpha)$.

$B_f$ defines a joint probability over $\{x, y, \mathbf{z}\}$ equivalent to model averaging over all sets of parameters for $D_f$. $\overline{N}_{XY\mathbf{Z}}$ is thus a Bayesian estimate of the expected sufficient statistics given the database $D$. Our independence test then uses this smoothed table to perform a classical hypothesis test; thus we call this a "pseudo-Bayesian" method. We label the algorithm corresponding to the PC algorithm using $Hybrid\text{-}IT(X, Y \mid \mathbf{Z}, D)$ the $PC^*$ *algorithm*. The time and space complexity required to calculate all parameters in $D_f$ are $O(N \cdot |N_{XY\mathbf{Z}}|)$, which is the same as that required to calculate $N_{XY\mathbf{Z}}$ directly from data.

The process of replacing the observed statistics of $D$ with the expected statistics defined by $B_f$ has the net result of changing the hypothesis being tested by $Std\text{-}IT(X, Y \mid \mathbf{Z}, \alpha)$. A standard hypothesis test tests whether the cells of the contingency table calculated by the data differ from the cells assuming independence; whereas $Hybrid\text{-}IT(X, Y \mid \mathbf{Z}, D)$ tests the same condition of the contingency table $\overline{N}_{XY\mathbf{Z}}$.

However, by projecting the distribution to be considered onto a particular Bayesian network structure, we have imposed independencies on $\overline{N}_{XY\mathbf{Z}}$, and it is not obvious that an independence in the smoothed contingency table gives us any information about the true independence of $X$ and $Y$ given the set $\mathbf{Z}$. The following theorems show that the process of constructing $D_f$ in Procedure 4, despite the fact that independencies between the $Z_i$ variables are imposed, will not alter the outcome of a perfect independence test $Std\text{-}IT(X, Y \mid \mathbf{Z}, N_{XY\mathbf{Z}})$.

The following is a well-known theorem, reproduced here for completeness:

**Lemma 1** *Let $P(\mathbf{V})$ be the joint distribution on the set $\mathbf{V} = \{X, Y\} \cup \mathbf{Z}$ for some set $\mathbf{Z} = \{Z_1, \ldots, Z_n\}$, then for arbitrary $x_i, x_j \in Rng(X)$, $y \in Rng(Y)$ and* $z \in Rng(\mathbf{Z})$, $P(y \mid \mathbf{z}, x_i) = P(y \mid \mathbf{z}, x_j)$ if and only if $(X \perp Y \mid \mathbf{Z})$.

**Definition 2 (projection)** *If $P(\mathbf{V})$ is a joint probability distribution over a set of variables $\mathbf{V}$, and $G(\mathbf{V})$ is a DAG over $\mathbf{V}$ then the projection $P(\mathbf{V})_G$ of $P$ onto $G$ is the distribution defined by the Bayesian network $\langle G, \boldsymbol{\theta} \rangle$, where the parameters $\boldsymbol{\theta}$ are given by the associated conditional distributions in $P$:*

$$\theta_{ijk} = P(X_i = x_i^k \mid \mathbf{P}_i = \mathbf{p}_i^j), \forall \theta_{ijk} \in \boldsymbol{\theta}$$

**Theorem 1** *Let $P(\mathbf{V})$ be the joint distribution on the set $\mathbf{V} = \{X, Y\} \cup \mathbf{Z}$ for some set $\mathbf{Z} = \{Z_1, \ldots, Z_n\}$, and let $G$ be the graph fragment defined in Procedure 4 for the test $(X \perp Y \mid \mathbf{Z})$. Then*

$$(X \perp Y \mid \mathbf{Z})_P \Leftrightarrow (X \perp Y \mid \mathbf{Z})_{P_G}.$$

**Proof:** For arbitrary $\mathbf{v} = \{x, y\} \cup \mathbf{z}$ such that $x \in Rng(X)$, $y \in Rng(Y)$, and $\mathbf{z} \in Rng(\mathbf{Z})$, by definition of $P(\mathbf{V})_G$, $P(y \mid x, \mathbf{z})_G = P(y \mid x, \mathbf{z})$; the result follows from Lemma 1. □

Theorem 1 shows that the structure of $G_f$ defined by Procedure 4 does not impose constraints that will alter the test of conditional independence $(X \perp Y \mid \mathbf{Z})$.

Using *Hybrid-IT*$(X, Y \mid \mathbf{Z}, D)$ has several advantages over a standard independence test: First, the use of Equation 2 allows prior knowledge to be accounted for in a normative fashion (although priors over structure cannot be specified explicitly, Heckerman et al. [8] suggest a means of deriving hyperparameters $\alpha_{ijk}$ that are consistent with a prior network elicited from an expert). Second, calculating parameters using Equation 2 can be accomplished even when the database $D$ is incomplete by using the EM algorithm or MCMC methods. The EM algorithm requires Bayesian network inference to be performed, so can be slow in general; however, due to the fact that the network fragments are small, in principle it should be relatively fast for this particular task. Finally, the use of even non-informative priors should provide a smoothing effect which improves the quality and stability of independence tests, especially for high-order tests and small data sets. In Section 4 we demonstrate these benefits empirically.

## 4 Experimental Tests

In this section we describe experimental investigations that were designed to test the performance of Procedure 4 on synthetic data.

For all experiments networks were generated randomly using the following procedure which directed arcs from $X_i \to X_j$ only if $j > i$:

**Procedure 5 (Random structure generation)**
**Given:** $N$ and $K$.
Do:

1. Create $N$ nodes $X_1, X_2, \ldots, X_N$.

2. For each node $X_i$ do:
   (a) Let $N_{pmax}^i \equiv \min(i-1, K)$.
   (b) Generate a random integer $N_{pa} \in [0, N_{pmax}^i]$.
   (c) Randomly pick $N_{pa}$ parents uniformly from the list $\{X_1, \ldots, X_{i-1}\}$

Once a network structure had been generated, each node distribution $\theta_{ij}$ was sampled independently from a uniform distribution over parameters. In all experiments we assumed the generating graph was sparse, i.e., $K = 5$.

We tested PC* against PC with a standard contingency table and against a Bayesian search procedure that used a greedy thick-thin (GTT) approach. GTT starts with an empty graph and repeatedly adds the arc (without creating a cycle) that maximally increases the marginal likelihood $P(D \mid S)$ until no arc addition will result in a positive increase, then it repeatedly removes arcs until no arc deletion will result in a positive increase in $P(D \mid S)$.

The outer-loop of each test performed the same procedure: Given the number of nodes $N$, number of records $N_r$ and total number of trials $N_{trials}$, a method $M$ was compared to PC* by doing the following:

**Procedure 6 (Basic testing loop)**
**Given:** $N$, $N_r$, and $N_{trials}$. Do:

1. Generate $N_{trials}$ random Bayesian networks $\hat{B}(N)$.

2. For each network $\hat{B}(N)$ do:
   (a) Generate $N_r$ records.
   (b) Learn a pattern $P$ with $M$, and learn pattern $P^*$ with $PC^*$.
   (c) Generate DAGs $G$ and $G^*$ by randomly directing all undirected edges of $P$ and $P^*$, respectively, without creating new v-structures or cycles.
   (d) Construct the Bayesian networks $B = \langle G, \boldsymbol{\theta} \rangle$ and $B^* = \langle G^*, \boldsymbol{\theta}^* \rangle$ using Equation 2 to calculate $\boldsymbol{\theta}$ and $\boldsymbol{\theta}^*$.
   (e) Measure the number of incorrect adjacencies $A$, $A^*$ and incorrect v-structures $V$, $V^*$ for $P$ and $P^*$, respectively.
   (f) Calculate the differences $\Delta_{adj} = A - A^*$ and $\Delta_v = V - V^*$ between the number of incorrect adjacencies and v-structures, respectively:

(g) Calculate the percent increase $\Delta_{kl}$ in KL-Divergence between $B$ and $B^*$:

$$\Delta_{kl} = \frac{KL\text{-}Div(\hat{B}, B) - KL\text{-}Div(\hat{B}, B^*)}{KL\text{-}Div(\hat{B}, B^*)}$$

3. Average $\Delta_{adj}$, $\Delta_v$, and $\Delta_{kl}$ over all $N_{trials}$.

Some final experimental details: (1) The running times $\tau$ and $\tau^*$ of each algorithm was recorded. (2) In all experiments we adopted the K2 criterion [3] which sets $\alpha_{ijk} = 1$ for all $(i, j, k)$. This criterion has the property of weighting all distributions of parameters equally. (3) All variables in our tests were binary: $r_i = 2$ for all $i$. (4) Except when explicitly mentioned, we used a significance level of $\alpha = 0.05$ for our statistical tests. All code used for our experiments was based on SMILE [5], a C++ library for constructing probabilistic decision support models.[1]

## 4.1 Experiments

Our experiments performed Procedure 6 with $N \in \{10, 20, 40, 80\}$ for a range of number of records $N_r$. For $N \in \{10, 20\}$, $N_{trials} = 1000$; for $N \in \{40, 80\}$, $N_{trials} = 100$. The results showing the performance of PC* over PC and GTT are shown in Table 1. The columns labelled CI denote the one-sided 99% confidence intervals. Positive results in any column indicates that PC* recorded fewer structural mistakes or lower KL-divergence.

These results show that PC* constructs significantly better networks by all three measures at low $N_r$ than does PC; however, as $N_r$ increases the difference decreases, sometimes losing the 99% significance. In only six of the 54 measurements did PC outperform PC* at the 99% level. These results also show that at low $N_r$ and high $N$ PC* outperforms GTT in terms of both structural features and KL-divergence. However, as the number of records per node increases, GTT begins to make significant gains on PC*; however, it was interesting to note that even as $N_r$ increased to its highest measured values, PC* typically made fewer errors (significant at the 99% level) in terms of the adjacencies of the network than did GTT.

One reason for the dramatic improvement in PC* over PC as $N$ is increased or $N_r$ is decreased is due to a known problem of CB algorithms. As $N$ grows large or $N_r$ grows small, PC has a tendency to *stall*, i.e., the time $t^l_{PC}$ for a particular run $l$ to finish could be much greater than the average time $\tau_{PC}$ to finish. This is due to the ironic fact that if the data is small the $chi\text{-}squared(X, Y \mid \mathbf{Z}, \mathbf{N}_{XYZ})$ test with a non-smoothed table will be more likely to *discover* independence relations because of noise in the tables. This in turn causes the PC algorithm to remove edges in the network that are critical to establishing separations later in the search, causing an overall more dense structure. Thus as the parameter $n$ in the PC algorithm increases, the average clique sizes in the network increase, causing an exponential increase in the number of conditioning sets to check. In these cases the end result was that PC would get caught in an intractable calculation that could not be finished in any reasonable time. The way this is handled in practice is that PC is set to exit whenever the conditioning set becomes larger than some integer $z$, or when some maximum time has elapsed.

We analyzed how the use of our hybrid independence test affected the frequency of the PC algorithm stalling. To this end, it was necessary to establish a cutoff time $\tau$ after which it was assumed that the PC algorithm had stalled. The following observations allowed the cutoff time to be established based on the time $t^*$ of the PC* algorithm:

1. The PC* algorithm rarely stalled.

2. A single conditional independence test using Hybrid-IT$(X, Y \mid \mathbf{Z}, D)$ typically took between 2 to 10 times longer than a chi-squared$(X, Y \mid \mathbf{Z}, \mathbf{N}_{XY\mathbf{Z}})$ test.

3. Aside from the independence test being used, PC and PC* are identical algorithms and were being tested on identical data sets.

---

[1]SMILE can be downloaded from http://www2.sis.pitt.edu/~genie; however the learning functionality required for our experiments is not yet available for public release.

|  |  | $N_r$ | $\Delta_{Adj}$ | CI | $\Delta_v$ | CI | $\Delta_{kl}$ | CI |
|---|---|---|---|---|---|---|---|---|
| v.s. PC | N=10 | 50 | 1.63 | 0.24 | 8.07 | 0.95 | 21.8% | 3.6% |
| | | 100 | 0.81 | 0.19 | 4.21 | 0.71 | 12.7% | 3.1% |
| | | 400 | 0.22 | 0.10 | 1.15 | 0.36 | 6.1% | 2.5% |
| | | 1600 | 0.05 | 0.05 | 0.14 | 0.15 | 5.7% | 4.0% |
| | | 6400 | -0.06 | 0.04 | -0.08 | 0.09 | 110% | 320% |
| | N=20 | 50 | 8.44 | 0.66 | 55.9 | 5.6 | 29.9% | 2.6% |
| | | 100 | 3.86 | 0.48 | 26.2 | 4.1 | 15.9% | 2.5% |
| | | 400 | 0.83 | 0.22 | 5.6 | 1.8 | 4.6% | 2.0% |
| | | 1600 | 0.28 | 0.13 | 1.6 | 1.1 | 3.2% | 1.8% |
| | | 6400 | -0.14 | 0.07 | -0.60 | 0.42 | 3.5% | 2.5% |
| | N=40 | 100 | 9.1 | 2.6 | 84 | 42 | 14% | 10% |
| | | 400 | 2.4 | 1.4 | 43 | 31 | 5.8% | 8.1% |
| | | 1600 | -0.5 | -1.0 | 2 | 12 | 8.1% | 6.1% |
| | | 6400 | -1.9 | 0.6 | -10.8 | 2.3 | 5.2% | 9.0% |
| | N=80 | 200 | 20.1 | 5.3 | 530 | 240 | 32% | 16% |
| | | 400 | 6.5 | 2.7 | 160 | 110 | 35% | 23% |
| | | 1600 | -2.1 | -1.0 | -4 | 7.8 | 8.4% | 7.2% |
| | | 6400 | -2.7 | 3.8 | 60 | 150 | 12% | 15% |
| v.s. GTT | N=10 | 50 | 1.90 | 0.24 | 4.65 | 0.23 | 19.4% | 1.8% |
| | | 100 | 0.88 | 0.22 | 2.27 | 0.19 | 0.3% | 1.1% |
| | | 400 | -0.06 | 0.19 | -0.33 | 0.27 | -12.5% | 0.9% |
| | | 1600 | 0.61 | 0.19 | -0.80 | 0.35 | -12.2% | 0.8% |
| | | 6400 | 1.65 | 0.20 | -0.18 | 0.39 | -9.9% | 0.7% |
| | N=20 | 50 | 16.47 | 0.43 | 30.25 | 0.68 | 190.9% | 4.8% |
| | | 100 | 8.78 | 0.40 | 13.16 | 0.43 | 46.1% | 2.6% |
| | | 400 | 0.88 | 0.36 | -2.17 | 0.50 | -36.2% | 1.6% |
| | | 1600 | 0.78 | 0.36 | -8.64 | 0.71 | -43.0% | 1.6% |
| | | 6400 | 4.01 | 0.35 | -9.07 | 0.87 | -37.1% | 1.7% |
| | N=40 | 100 | 62.1 | 3.5 | 117 | 38 | 30.5% | 9.2% |
| | | 400 | 11.7 | 2.7 | 7 | 26 | -43.9% | 4.3% |
| | | 1600 | -1.1 | 1.7 | -30.2 | 5.9 | -80.2% | 1.8% |
| | | 6400 | 5.4 | 2.0 | -36.7 | 6.1 | -90.9% | 1.4% |
| | N=80 | 200 | 163.0 | 6.7 | 260 | 160 | 16% | 10% |
| | | 400 | 83.6 | 5.8 | 171 | 77 | -20.1% | 4.8% |
| | | 1600 | 1.8 | 3.9 | -92 | 65 | -78.7% | 1.3% |
| | | 6400 | 7.6 | 2.7 | -97.7 | 9.4 | -91.0% | 1.0% |

Table 1: The performance of PC* over PC and GTT as $N_r$ is varied for $N = 10$ and $N = 20$.

These facts allowed a reasonable cutoff time for PC to be tied to the longer but more reliable exit time $t^*$ of the PC* algorithm. For example, the criterion $\tau = t^*$ would not have been unreasonable since we expect PC* to take 2–10 times longer than PC. In fact, we used a much more conservative criterion, choosing $\tau$ to be greater than 5 standard deviations from $t^*$, i.e., such that the probability $P(t^* < \tau) > 1 - 10^{-6}$. This procedure made possible measurements in high $N$, low $N_r$ regimes where the PC algorithm will stall a majority of the time.

Figure 1 shows the probability of stall for both the PC and PC* algorithms for several configurations of $\{N, N_r\}$. It is evident from this figure that the Hybrid-IT$(X, Y \mid \mathbf{Z}, D)$ all but eliminates the probability of a stall for all tested values of $N$ and $N_r$. This is sharply contrasted to the standard implementation of the chi-squared$(X, Y \mid \mathbf{Z}, \mathbf{N}_{XY\mathbf{Z}})$ test which for large $N$ and small $N_r$ can achieve stall probabilities approaching 100%.

The comparisons between PC and PC* were vulnerable to the criticism that the difference in KL-Divergence between PC and PC* might have been due to the fact that our selected significance level $\alpha = 0.05$ for some reason favored the Hybrid-IT$(X, Y \mid \mathbf{Z}, D)$ over the standard chi-squared$(X, Y \mid \mathbf{Z}, \mathbf{N}_{XY\mathbf{Z}})$ test. A less naive experiment would have first tuned $\alpha$ for PC then separately tuned $\alpha$ for PC* for the comparisons. Thus if we just so happened to pick an $\alpha$ that benefited PC*, that could explain our experimental results.

To test this hypothesis we performed a test with $N = 10$, $N_r = 100$ and with $\alpha$ varied over more than three decades from 0.0001 to 0.2. If it was possible to tune $\alpha$ to PC in such a way that it performed better than PC*, then our results would be in doubt. These results are shown in Figure 2. Figure 2(a) shows the value of $\Delta_{kl}$ as $\alpha$ is varied; whereas Figure 2(b) shows the overall KL-Divergence of PC as $\alpha$ is varied. It was observed that as $\alpha$ was reduced to extremely small values the value of $\Delta_{kl}$ began to approach zero; however, for these values of $\alpha$ the overall quality of PC graphs decreased as indicated by the increasing KL divergence. Conversely, as the quality of PC graphs was tuned to its optimum value the value of $\Delta_{kl}$ achieved a maximum. This experiment demonstrated for at least the $\{N = 10, N_r = 100\}$ configuration that the gains in $\Delta_{kl}$ shown in Table 1 were not due to a shifting in the optimum significance level between PC and PC*.

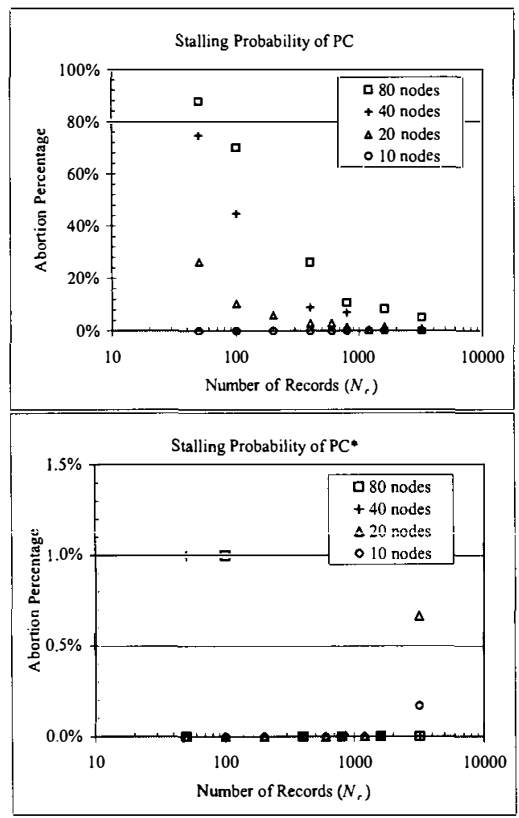

Figure 1: Probabilities of the PC and PC* algorithms stalling.

## 5　Conclusions

We have demonstrated that a Hybrid independence test can be used along with a set of (possibly non-informative) priors to produce more robust independence tests. We have demonstrated empirically that using PC* consistently decreases the KL-Divergence of networks compared to PC, recovers structural features more accurately and dramatically reduces the probability of PC getting stuck on small data sets. The improvements to PC were significant enough for it to outperform a greedy algorithm based on Bayesian techniques when the database is small or when the number of nodes increases.

This technique is easy to implement: any exist-ing CB algorithm can be modified simply by replacing the independence test and leaving the rest of the algorithm untouched. The benefits of using $Hybrid\text{-}IT(X, Y \mid \mathbf{Z}, D)$ were not without cost. Typically the time taken to learn a graph was $2-5$ times longer for PC* than for the PC runs that did not stall.

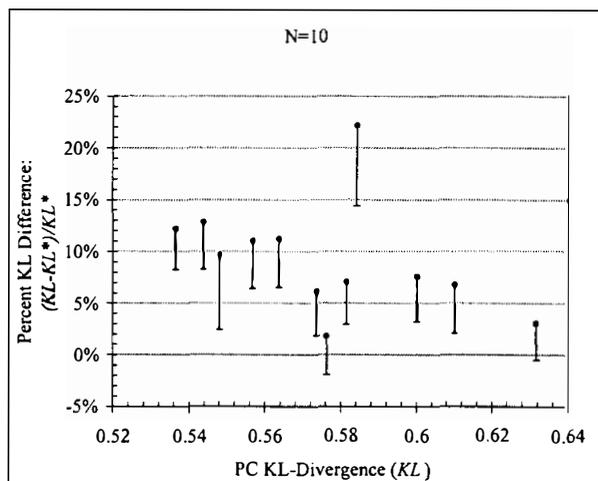

Figure 2: The relative performance of PC* as the significance level $\alpha$ is tuned for PC. As the KL-divergence for PC is minimized, the performance in PC* increases. The error bars denote the one-sided 99% confidence intervals.

This test allows CB learning to be performed even when the database is partially missing data using the EM algorithm or MCMC methods. The ability to learn structure with missing data opens up the possibility of using CB techniques for unsupervised classification. It is interesting to see how a CB unsupervised classifier will perform compared to one learned using Bayesian methods.

## 6 Acknowledgements

This work was supported in part by the National Aeronautics and Space Administration under the Graduate Students Research Program grant S99-GSRP-085 and by the Air Force Office of Scientific Research grant F49620-00-1-0112.

## References

[1] David Maxwell Chickering. Learning equivalence classes of Bayesian network structures. In *Proceedings of the Twelfth Annual Conference on Uncertainty in Artificial Intelligence (UAI-96)*, pages 150–157, San Francisco, CA, 1996. Morgan Kaufmann Publishers.

[2] Greg Cooper. A simple constraint-based algorithm for efficiently minining observational databases for causal relationships. *Data Mining and Knowledge Discovery*, 1:203–224, 1997.

[3] Gregory F. Cooper and Edward Herskovits. A Bayesian method for the induction of probabilistic networks from data. *Machine Learning*, 9(4):309–347, 1992.

[4] Denver Dash and Marek Druzdzel. A hybrid anytime algorithm for the construction of causal models from sparse data. In *Proceedings of the Fifteenth Annual Conference on Uncertainty in Artificial Intelligence (UAI-99)*, pages 142–149, San Francisco, CA, 1999. Morgan Kaufmann Publishers.

[5] Marek J. Druzdzel. SMILE: Structural Modeling, Inference, and Learning Engine and GeNIe: A development environment for graphical decision-theoretic models. In *Proceedings of the Sixteenth National Conference on Artificial Intelligence (AAAI-99)*, pages 902–903, Orlando, FL, July 18–22 1999.

[6] Nir Friedman, Iftach Nachman, and Dana Pe'er. Learning Bayesian network structure from massive datasets: The "sparse candidate" algorithm. In *Proceedings of the Fifteenth Annual Conference on Uncertainty in Artificial Intelligence (UAI-99)*, pages 196–205, San Francisco, CA, 1999. Morgan Kaufmann Publishers.

[7] David Heckerman. A tutorial on learning with Bayesian networks. In Michael I. Jordan, editor, *Learning in Graphical Models*. The MIT Press, Cambridge, Massachusetts, 1998.

[8] David Heckerman, Dan Geiger, and David M. Chickering. Learning Bayesian networks: The combination of knowledge and statistical data. *Machine Learning*, 20:197–243, 1995.

[9] David Heckerman, Christopher Meek, and Gregory F. Cooper. A bayesian approach to causal discovery. In *Computation, Causation, and Discovery*, chapter four, pages 141–165. 1999.

[10] Tomas Kocka and Robert Castelo. Improved learning of bayesian networks. In *Uncertainty in Artificial Intelligence: Proceedings of the Seventeenth Conference (UAI-2001)*, pages 269–276, San Francisco, CA, 2001. Morgan Kaufmann Publishers.

[11] David Madigan, Steen A. Anderson, Michael D. Perlman, and Chris T. Volinsky. Bayesian model averaging and model selection for Markov equivalence classes of acyclic digraphs. *Communications in Statistics—Theory and Methods*, 25(11):2493–2519, 1996.

[12] Moninder Singh and Marco Valtorta. An algorithm for the construction of Bayesian network structures from data. In *Proceedings of the Ninth Annual Conference on Uncertainty in Artificial Intelligence (UAI-93)*, pages 259–265, San Francisco, CA, 1993. Morgan Kaufmann Publishers.

[13] Peter Spirtes, Clark Glymour, and Richard Scheines. *Causation, Prediction, and Search*. Springer Verlag, New York, 1993.

[14] Peter Spirtes and Christopher Meek. Learning Bayesian networks with discrete variables from data. In *Proceedings of the First International Conference on Knowledge Discovery and Data Mining*, pages 294–299, 1995.